\documentclass{article}
\usepackage{ijcai18}

\usepackage{times}
\usepackage{xcolor}
\usepackage{soul}
\usepackage[utf8]{inputenc}
\usepackage[small]{caption}

\usepackage{times}
\usepackage{epsfig}
\usepackage{graphicx}
\usepackage{amsmath}
\usepackage{amssymb}
\usepackage{subfigure}
\usepackage{array}
\usepackage{booktabs}
\usepackage{setspace}
\usepackage{epstopdf}

\title{End-to-End Detection and Re-identification Integrated Net for Person Search}

\author{
Zhenwei He$^1$,
Lei Zhang$^1$$^*$,
Wei Jia$^2$
\\
$^1$ College of Communication Engineering, Chongqing University\\
$^2$ School of Computer and Information, Hefei University of Technology  \\
hzw@cqu.edu.cn,
leizhang@cqu.edu.cn,
china.jiawei@139.com
}

\begin{document}

\maketitle

\begin{abstract}
This paper proposes a pedestrian detection and re-identification (re-id) integration net (I-Net) in an end-to-end learning framework. The I-Net is used in real-world video surveillance scenarios, where the target person needs to be searched in the whole scene videos, while the annotations of pedestrian bounding boxes are unavailable. By comparing to the successful CVPR'17 work \cite{Xiao_2017_CVPR} for joint detection and re-id, we have three distinct contributions. First, we introduce a Siamese architecture of I-Net instead of 1 stream, such that a verification task can be implemented. Second, we propose a novel on-line pairing loss (OLP) and hard example priority softmax loss (HEP), such that only the hard negatives are posed much attention in loss computation. Third, an on-line dictionary for negative samples storage is designed in I-Net without recording the positive samples. We show our result on person search datasets, the gap between detection and re-identification is narrowed. The superior performance can be achieved.
\end{abstract}

\section{Introduction}

Real-world video surveillance tasks such as criminals search~\cite{Wang2013Intelligent}, multi-camera tracking~\cite{Song2010Tracking} need to search the target person from different scenes. Additionally, in real-world person search tasks, the algorithms are asked to find the target person from whole image scene. Therefore, this problem is generally issued by two separate steps: person detection from an image and person re-identification (re-id). These two problems are challenging due to the influences of poses, viewpoints, lighting, occlusion, resolution, background \emph{etc}. Therefore, these two problems have been paid too much attention~\cite{Chen2016Deep,Cao2017Solving,Song2017Collaborative,Yang2017Enhancing}.

Although numerous endeavor on person detection and re-identification has been made, most of them handle the two problems independently. The traditional methods for person search task generally divide the task into two sub-problems. First, a detector is trained to predict the bounding boxes of persons from the image scene. Second, the persons are cropped based on the bounding boxes, which are used to train a re-id identifier for target person matching. Actually, most advanced re-identification method are modeled on the manually cropped pedestrian images~\cite{Chen2016Deep,Gray2007Evaluating}, and the cropped pedestrian samples are much better than the specially trained detector because of inevitable false detection. Additionally, it does not comply with the real-world person search application that person search task should be a joint work of detection and re-id, instead of separate ones. Also, person search task requires the close cooperation of the detector and the identifier. Therefore, in this paper, we propose to jointly modeling these two parts in a unified deep framework by end-to-end learning. Our ultimate goal is to search a target person from the whole image scene directly without cropping images.

\begin{figure}[t]
\begin{center}
   \subfigure[Traditional method]{
   \label{Figure 1.a}
   \includegraphics[width=0.9\linewidth]{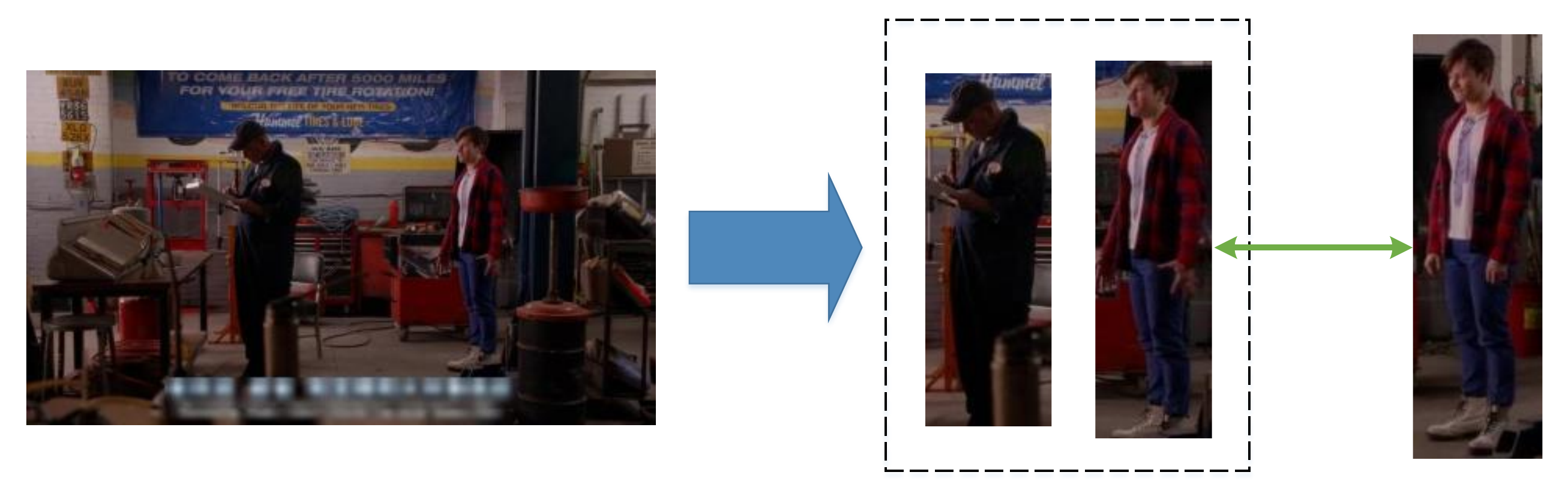}
   }
   \hspace{1in}
   \subfigure[Our method]{
   \label{Figure 1.b}
   \includegraphics[width=0.9\linewidth]{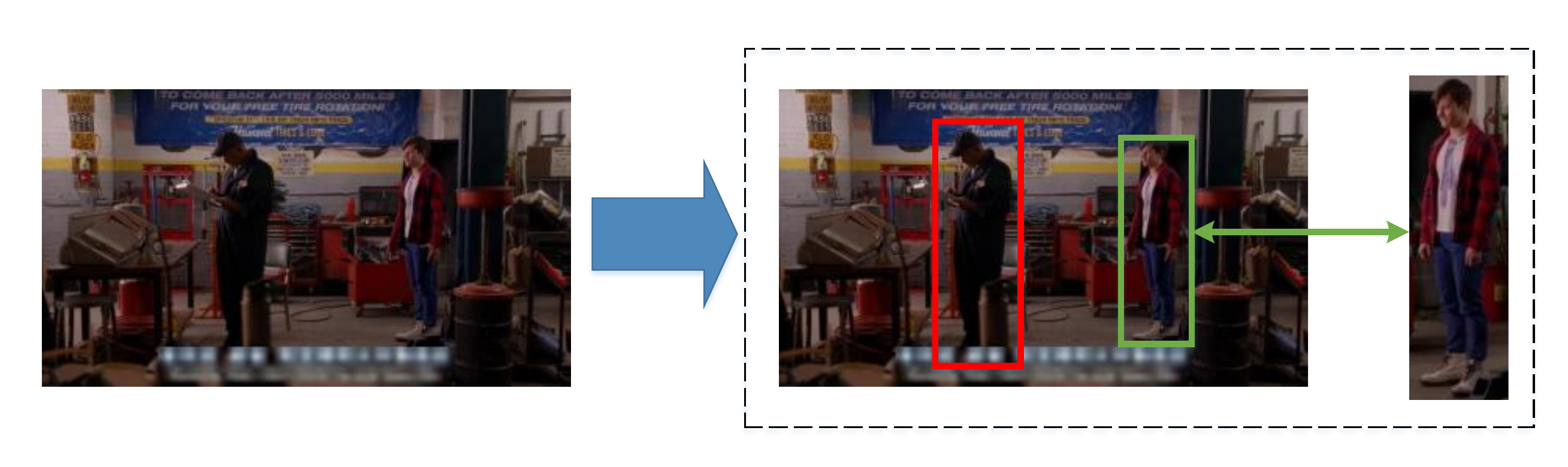}
   }
\end{center}
   \caption{Compare of the traditional method and our new method. In the Figure 1.a, Traditional method need to detect all persons in the picture and cropped it to extract futures for further re-id model, which stiffly slice the person search task into two sub-problems. Our new method(I-Net) in Figure 1.b can find the target person from a whole image directly without cropping the images.}
\label{fig:compare}
\end{figure}

Specifically, to close the gap between traditional algorithms and practical applications, we propose an Integration Net (I-Net) to simultaneously learn the detector and re-identifier for person search by an end-to-end manner. Fig.\ref{fig:compare} shows the difference of application scenario between traditional scheme and the proposed I-Net program. Different from the traditional re-id method, our I-Net can predict the location (bounding box) of the target person directly. The joint learning of the detector and re-identifier in I-Net for person search brings lots of benefits. On one hand, the co-learning of the detector and re-identifier helps to handle misalignments, such that the re-id part can be more robust than independent training. On the other hand, the detection and re-id share the same feature representation which can reduce the computation time and accelerate the search speed. In the proposed I-Net, for detection and re-id, VGG16~\cite{Simonyan2014Very} network is shared for feature representation.

Further, in order to achieve the purpose of co-learning of person detection and re-identification, a novel on-line pairing loss (OLP loss) and a hard example priority softmax loss (HEP loss) are proposed in I-Net. By storing the features from different persons in a dynamic dictionary, a positive pair and lots of negative pairs can be captured during the training phase. OLP loss calculates a cosine distance based on softmax function, in which a symmetric pairing of anchor is proposed. In OLP loss, a mass of negative pairs can constrain the condition of the positive pair more strict. HEP loss is an auxiliary loss function based on softmax loss, which is calculated by considering only the hard examples with high priority. Different from the OIM loss in~\cite{Xiao_2017_CVPR}, by unifying the whole process into a Siamese architecture, one real-time positive pair and lots of negative pairs can be obtained in the dynamic dictionary. With the specially designed loss functions and training strategy, the proposed I-Net demonstrates a good efficiency and effectiveness.

\section{Related Work}

Person search task need 2 aspects works:detection and re-identification, our work is based on both of them. For re-id problem, our OLP loss is inspired by triplet loss~\cite{Schroff2015FaceNet}, which is widely used in person re-identification~\cite{Cheng2016Person,Liu2016Multi} in recent years. Due to the limited number of negative pairs, the positive pairs can easily satisfy the loss function, such that the training phase is stagnate. In order to deal with this issue, we store amounts of features to extend the number of negative pairs in OLP loss function.

On the other hand, detection is another important issue. Traditional pedestrian detection methods are based on hand-crafted features and Adaboost classifiers, such as ACF~\cite{Dollar2014Fast}, LDCF~\cite{Nam2014Local} and Integral Channels Features (ICF) \cite{Doll2009Integral}. Recently, convolutional neural network (CNN) based deep learning methods have achieved significant progress in detection. Our method is based on the Faster-RCNN~\cite{Ren2017Faster}, which is jointly training a region proposal network (RPN) that shares the model with VGG16~\cite{Simonyan2014Very} network in feature representation. We adapt the RPN into our I-net for further re-id task.

Recently, some person search methods are proposed, such as OIM~\cite{Xiao_2017_CVPR} and NPSM~\cite{Liu2017Neural}. OIM stores features of each id in order to train the model while NPSM searches the regions contain the target person from whole image. However, the features OIM stored aren't update in time and the computation cast of NPSM is very high. Our method overcome these disadvantages and achieves state-of-art result.

\section{The Proposed I-Net}

\begin{figure*}
\begin{center}
   \includegraphics[width=0.9\linewidth]{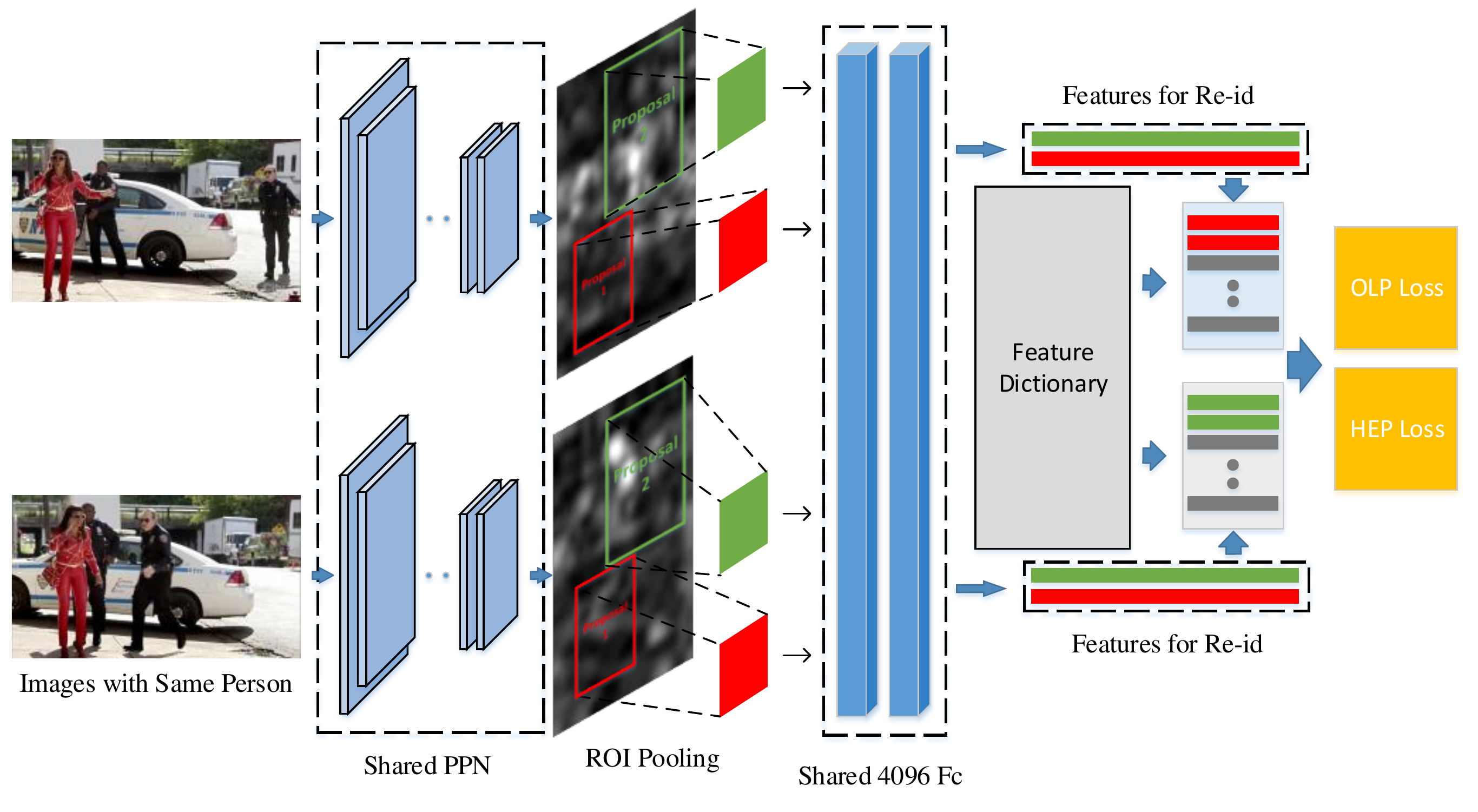}
\end{center}
   \caption{The net structure of our I-Net. A pair of images with same identity are feed into our network. Two PPNs shared same parameters are used to generate proposals for pedestrians from images. These proposals are feed into the further fc layers to get the corresponding features. The input features combined with feature dictionary are then used to form positive and negative pairs, which participate the computation of OLP loss and HEP loss.}
\label{fig:netstructure}
\end{figure*}

We propose a new I-Net framework that jointly handles the pedestrian detection and person re-identification in an end-to-end Siamese network. The architecture is shown in Fig.\ref{fig:netstructure}. Given a pair of images with persons of the same identity, two pedestrian proposal networks (PPN) with shared parameters are learnt to predict the proposals for pedestrians. The feature maps pooled by ROI pooling layer are then feed into the fully-connected (fc) layers to extract 256-D L2-normalized features. These features are then stored in an on-line dynamic dictionary where one positive pair and lots of negative pairs are generated for computation of OLP loss and HEP loss.

\subsection{Deep Model Structure}

The basic model of I-Net is based on the VGG16 architecture~\cite{Simonyan2014Very}, which has 5 stacks of convolutional part, including 2, 2, 3, 3, 3 convolutional layers for each stack. 4 max-pooling layer are followed on the first 4 stacks. In the $conv5\_3$ layer, it generates 512 channels for features of the PPN outputs. These feature maps have 1$ \verb|\|$16 resolutions of the original image after down-sampling by the 4 max-pooling layers.

On the top of $conv5\_3$ feature map, PPN is used to generate pedestrian proposals. A $512\times3\times3$ convolutional layer is first added to get the features for pedestrian proposals. Similar to faster RCNN \cite{Ren2017Faster}, we then associate 9 anchors at each feature map, and a softmax classifier (cls.) is used to predict whether the anchor is a pedestrian or not. A smooth L1 Loss (reg.) is used for pedestrian locations (bounding box) regression. Finally, 128 proposals for each image after the non-maximum suppression (NMS) are obtained.

A ROI pooling layer \cite{Girshick2015Fast} is integrated in I-Net to pool the generated proposals from the $conv5\_3$ feature map. The pooled feature is then feed into the 2 fc layers of 4096 neurons. In order to remove the false positives of the proposals, a 2 class softmax layer is trained to classify the proposals. Then a 256-D L2-normalized \cite{Liu2017SphereFace} feature is feed into the OLP loss and HEP loss for guiding the whole training phase of I-Net. Together with the loss function (cls vs. reg) of faster RCNN, the proposed I-Net can be jointly trained for simultaneous person detection and re-identification in an end-to-end architecture.

\subsection{On-line Pairing Loss (OLP)}

\begin{figure*}
\begin{center}
   \includegraphics[width=0.9\linewidth]{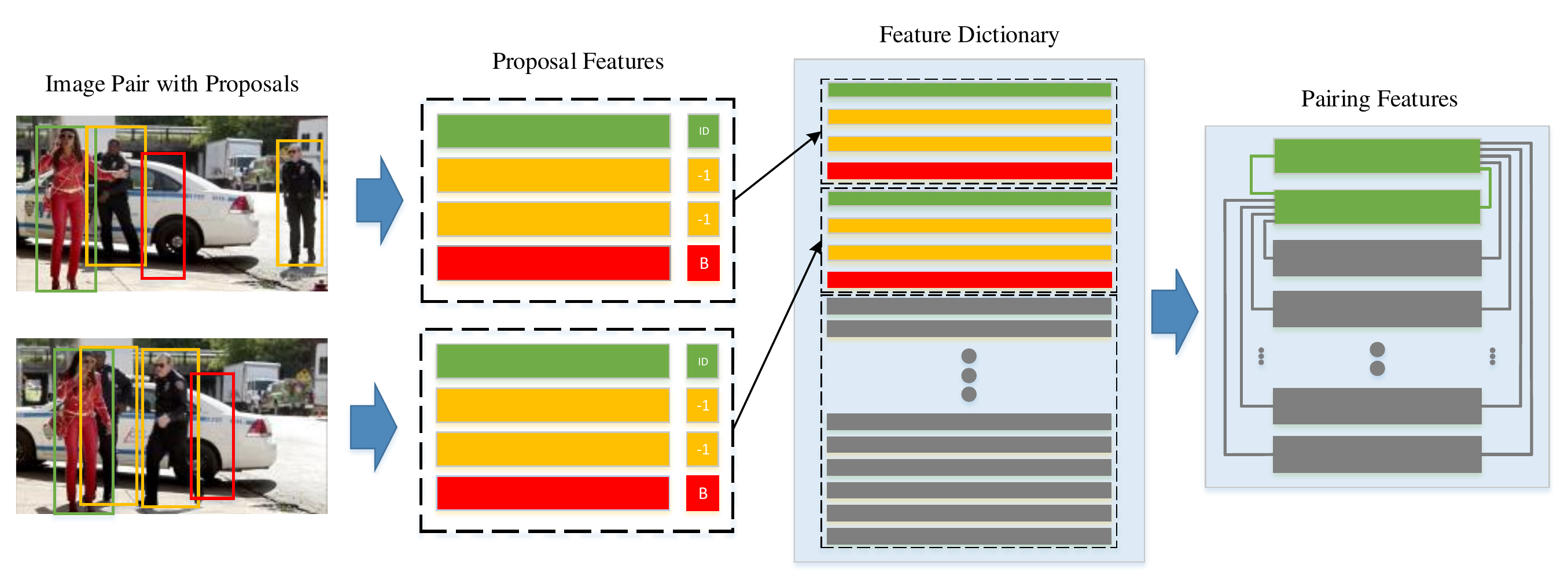}
\end{center}
   \caption{The work procedure of OLP. The feature of all types of proposals including background(red), person with id information(green) and person without id information(yellow and marked as -1) are extracted. These features are stored into the feature dictionary. OLP loss use the proposal of the same id person from input images as a positive pair(the two green feature at right of the picture collected by green lines) while the features in the feature dictionary (labeled with other id) is used to construct negative pairs (collected with gray lines).}
\label{fig:OLPloss}
\end{figure*}

In detection part, 128 proposals per image are learned, which are then feed into the re-identification part. For person re-id, the proposal features can be divided into 3 types, including background (B), persons with identity information (p-w-id) and persons without identity information (p-w/o-id). The division depends on the IOU between the proposals and ground truth. As shown in Fig.\ref{fig:OLPloss}, the background, p-w-id and p-w/o-id are represented by red, green, yellow bounding box respectively. In the OLP loss, an on-line feature dictionary is designed where the features of all person proposals and part of background proposals with their labels in an image are stored. Note that, the number of the stored features depends on the mini-batch size. Specifically, the stored feature number is 40 times the mini-batch size. Notably, once the number of features in the dictionary reaches the maximum number, the earliest feature will be replaced with the new one.

The goal of I-Net is to distinguish different persons. In order to minimize the discrepancy of proposal features of the same person, while maximize the discrepancy of different person proposal features, we use the person proposals of the same identity from the image pair and the feature dictionary to establish positive and negative pairs. Suppose that the proposal group for loss computation is $(\textbf{p}_{1}, \textbf{p}_{2}, \textbf{n}_{1}, \textbf{n}_{2} ... , \textbf{n}_{k})$, where $(\textbf{p}_{1}, \textbf{p}_{2})$ stands for the proposals of the same identity from the input image pair and $(\textbf{n}_{1}, \textbf{n}_{2} ... , \textbf{n}_{k})$ are the proposals stored in the dictionary. For each proposal group, we tend to formulate two symmetrical subgroups by taking $\textbf{p}_{1}$ and $\textbf{p}_{2}$ as anchor, alternatively. Similar to the triplet loss \cite{Schroff2015FaceNet}, when $\textbf{p}_{1}$ is regarded as anchor, the $(\textbf{p}_{1}, \textbf{p}_{2})$ is the positive pair, $(\textbf{p}_{1}, \textbf{n}_{1})$, $(\textbf{p}_{1}, \textbf{n}_{2})$, ... , $(\textbf{p}_{1}, \textbf{n}_{k})$ are negative pairs. Alternatively, when $\textbf{p}_{2}$ is regarded as anchor, $(\textbf{p}_{2}, \textbf{p}_{1})$ is the positive pair, $(\textbf{p}_{2}, \textbf{n}_{1})$, $(\textbf{p}_{2}, \textbf{n}_{2})$, ... , $(\textbf{p}_{2}, \textbf{n}_{k})$ are negative pairs. Consider the large amount of negative samples, that may make the number of triplet pairs too large, the OLP loss is established based on softmax function. Suppose we get $m$ subgroups in one iteration, and $\textbf{x}_{A}^{i}$, $\textbf{x}_{p}^{i}$, $(\textbf{x}_{n_{1}}^{i}, \textbf{x}_{n_{2}}^{i} ... , \textbf{x}_{n_{k}}^{i})$ stands for the anchor, positive and negative features of $i^{th}$ subgroup respectively, the OLP loss function is represented as follows.
\begin{equation}\label{OLP_Loss}
  L_{OLP} = -\frac{1}{m}
  \sum_{i=1}^{m}
  log_{}\frac{e^{d(\textbf{x}_{A}^{i},\textbf{x}_{p}^{i})}}
  {e^{d(\textbf{x}_{A}^{i},\textbf{x}_{p}^{i})}+\sum_{j=1}^k
  e^{d(\textbf{x}_{A}^{i},\textbf{x}_{n_{j}}^{i})}}
\end{equation}
where the function $d(\cdot)$ stands for the cosine distance of two features. Because these features are L2-normalized, the cosine distance can be directly computed by the inner product of each two features. In gradient computation, we only calculate the deviation from the anchor feature, which is different from triplet loss where the deviations of anchor, positive and negative features have been computed.

Then, the deviation of the OLP loss function with respect to $\textbf{x}_{A}^{i}$ for the $i^{th}$ subgroup can be calculated as
\begin{equation}\label{hatq}
  \frac{\partial L_{OLP}}{\partial \textbf{x}_{A}^{i}} = (q^{i}-1)\textbf{x}_{P}^{i}+\sum_{l=1}^k{(\hat{q}_{l}^{i}\textbf{x}_{n_{l}}^{i})}
\end{equation}
where $q^{i}$ and $\hat{q}_{l}^{i}$ are expressed in equation (\ref{q}) and (\ref{qi}).
\begin{equation}\label{q}
  q^{i} = \frac{e^{d(\textbf{x}_{A}^{i},\textbf{x}_{p}^{i})}}
  {e^{d(\textbf{x}_{A}^{i},\textbf{x}_{p}^{i})}+\sum_{j=1}^k
  e^{d(\textbf{x}_{A}^{i},\textbf{x}_{n_{j}}^{i})}}
\end{equation}

\begin{equation}\label{qi}
  \hat{q}_{l}^{i} = \frac{e^{d(\textbf{x}_{A}^{i},\textbf{x}_{n_{l}}^{i})}}
  {e^{d(\textbf{x}_{A}^{i},\textbf{x}_{p}^{i})}+\sum_{j=1}^k
  e^{d(\textbf{x}_{A}^{i},\textbf{x}_{n_{j}}^{i})}}, l=1,...,k
\end{equation}

Following the standard BP optimization in CNN, stochastic gradient descent (SGD) is adopted in the training phase of I-Net.

From the OLP loss (\ref{OLP_Loss}), we can see that a large number of negative features can be processed at one time by utilizing the cosine distance guided softmax function. The problem of a large number of triplet pairs is then efficiently solved. In terms of the softmax character, the cosine distance between anchor and positive samples tend to be maximized. The performance of person re-id is then improved.

\subsection{Hard Example Priority Softmax loss (HEP)}

A person matching problem is supposed in re-identification. The OLP loss function aims to constrain the cosine distance of positive pairs to be larger than the cosine distance of negative pairs. In OLP loss, the identity information of persons that is useful to supervise the training phase is not fully used. Therefore, we propose a HEP softmax loss function for person identity classification. The traditional softmax loss computes the output of all neurons (the number of classes), such that the computational complexity is too high if the number of classes is too large. Different from the traditional softmax loss, we propose a hard example priority (HEP) strategy, which focus on the part classes of hard negative pairs.

Suppose that there are $C$ identities, the HEP loss function aims to classify all proposals (except the persons without id) into $C+1$ classes ($C$ classes plus a background class). Specifically, for calculating the HEP loss, the hard negative samples with high priority should be first selected. To this edit, when a subgroup is paired by OLP loss and their cosine distance of each negative pair can be computed. Therefore, we find out the top 20 maximum distance of the negative pairs, and record their corresponding labels indexed from the feature dictionary. As a result, the $K$ classes are recognized as hard negative classes with high priority. Therefore, we will preferentially choose the priority class for HEP softmax loss computation. On the other hand, the labels of each proposal is collected to used as true labels of these proposals. Additionally, in order to keep the total number of the selected classes fixed, we also randomly choose an uncertain number of classes from the remaining classes, such that totally $M (M<<C+1)$ classes are selected to compute the final loss. Note that $M$ is a hyperparameter and set as 100 in experiments. Finally, the HEP softmax loss function is represented as
\begin{equation}\label{HEPLoss}
    L_{HEP} = -\frac{1}{n}
  \sum_{i=1}^{n}
  \sum_{j=1}^{M}
  \textbf{1}(label = j)log_{}\frac{e^{x_{j}^{i}}}
  {\sum_{k=1}^{M}
  e^{x_{k}^{i}}}
\end{equation}
where $x_{j}^{i}$ stands for the $i$-th proposal's score from the classifier and $j$ stands for the $j$-th class. Suppose that $\mathbf{L}$ stands for the pool of chosen classes, then the protocol for choosing the $M$ classes with hard example priority and randomness is summarized as below.

\begin{enumerate}
\item The label indexes of generated proposals from input image pair are first stored in the label pool $\mathbf{L}$.

\item For each subgroup, the label indexes of the top 20 negative pairs with the maximum distance are recorded. The chosen labels from all subgroups are then stored in the label pool $\mathbf{L}$.

\item If the size of pool $\mathbf{L}$ is still smaller than $M$ (a preset value), then we randomly generate the label indexes without repetition and stored in the label pool $\mathbf{L}$.
\end{enumerate}

This strategy ensures that the classes of hard samples are preferentially selected. That is, if a person with identity is hard to distinguish from others, then this person proposal feature must participate the HEP softmax loss computation in Eq.(\ref{HEPLoss}).

\section{Experiments}

To evaluate the effectiveness of our approach for joint person detection and re-identification, we conduct a number of experiments on the CUHK-SYSU dataset~\cite{Xiao2016End}. We first describe the experimental settings and baselines. Then we compare the proposed I-Net with the baselines solving the person search problem. Discussion of our model is presented at last.

\subsection{Experimental Setting and Data}

Our I-Net is implemented on Caffe~\cite{Jia2014Caffe} and py-faster-rcnn~\cite{Ren2017Faster} platform. VGG16~\cite{Simonyan2014Very} is the basic network of I-Net and the trained model \cite{Xiao2016End} is used for network initialization. The first 2 stacks of convolutional layers are frozen while training our net. The two streams of the I-Net share the same parameters of VGG16 for both initialization and training. The pedestrian proposal network (PPN) at each branch generates 128 proposals with a ratio of 1:3 for foreground and background. We randomly choose 5 background proposals per image which are stored in the feature dictionary for OLP loss computation. In I-Net, all loss functions are imposed the same loss weight. The learning rate is initialized to 0.001, and drops to 0.0001 in 50k iterations. Totally, 60k iterations are set to insure convergence.

The CUHK-SYSU dataset~\cite{Xiao2016End} has 18184 images from different scenes, which is specially developed for person search in the whole image, are used in our experiments. By following the same experimental setting as ~\cite{Xiao2016End},~\cite{Xiao_2017_CVPR}, 11206 images and 6978 images have been used for training and testing, respectively. There are 5532 identities for training while 2900 identities are used in testing phase. In I-Net, the image pairs are formulated based on the 5532 identities in the training set, and finally we get about 16000 image pairs for training. In each epoch, we shuffle their order. Additionally, we also mirror the images in training set to augment the training data.

For baseline comparisons, we select 3 pedestrian detection methods and 4 person re-id approaches, which then result in 12 baselines. The 3 detection methods, CCF~\cite{Yang2015Convolutional}, Faster-RCNN~\cite{Ren2017Faster} with resnet50~\cite{He2016Deep} and ACF~\cite{Dollar2014Fast}, are used for detecting pedestrians. All of them are trained or fine-tuned on the CUHK-SYSU dataset~\cite{Xiao2016End}. Additionally, the ground truths (manually labeled person bounding box) are recognized as the perfect detector.

For person re-identification task, we evaluate several famous re-id feature representations including DenseSIFT-ColorHist (DSIFT)~\cite{Zhao2013Unsupervised}, Bag of Words (BoW)~\cite{Zheng2015Scalable}, and Local Maximal Occurrence (LOMO)~\cite{Liao2015Person}. Each feature representation is measured by some specific distance metrics, including Euclidean, Cosine similarity, KISSME~\cite{Roth2012Large}, and XQDA~\cite{Liao2015Person}. The KISSME~\cite{Roth2012Large} and XQDA~\cite{Liao2015Person} are trained on the CUHK-SYSU dataset~\cite{Xiao2016End} in experiments.

The above methods address the person search problem in separate work. Further, we have compared with the OIM loss model~\cite{Xiao_2017_CVPR}, the end-to-end model~\cite{Xiao2016End} and NPSM ~\cite{Liu2017Neural}, which addressed the same real-world person search in the whole image by jointly learning detection and re-id. Following the protocol of the CUHK-SYSU dataset~\cite{Xiao2016End}, the gallery size is set as 100 in experiments if not specified. We implement the source code of OIM~\cite{Xiao_2017_CVPR} to get our result.

\subsection{Experimental Results}

\begin{table}
\label{Table:res}
\begin{center}
\caption{Comparisons between our framework and other methods}
\begin{tabular}{c|c c c c}
\hline
CMC top-1(\%) & CCF & ACF & CNN & GT \\
\hline\hline
DSIFT+Euclidean & 11.7 & 25.9 & 39.4 & 45.9\\
DSIFT+KISSME & 13.9 & 38.1 & 53.6 & 61.9\\
BoW+Cosine & 29.3 & 48.4 & 63.1 & 76.7\\
LOMO+XQDA & 57.1 & 63.0 & 74.8 & 78.3\\
\hline
Initialized model & - & - & 62.7 & - \\
OIM(w/o unlabeled) & - & - & 76.1 & 78.5 \\
OIM & - & - & 78.7 & 80.5 \\
NPSM & - & - & 81.2 & - \\
I-Net(w/o unlabeled) & - & - & 81.3 & 81.7 \\
I-Net & - & - & 81.5 & 82.0 \\
\hline
mAP(\%) & CCF & ACF & CNN & GT \\
\hline\hline
DSIFT+Euclidean & 11.3 & 21.7 & 34.5 & 41.1\\
DSIFT+KISSME & 13.4 & 32.3 & 47.8 & 56.2\\
BoW+Cosine & 26.9 & 42.4 & 56.9 & 62.5\\
LOMO+XQDA & 41.2 & 55.5 & 68.9 & 72.4\\
\hline
Initialized model & - & - & 55.7 & - \\
OIM(w/o unlabeled) & - & - & 72.7 & 75.5 \\
OIM & - & - & 75.5 & 77.9 \\
NPSM & - & - & 77.9 & - \\
I-Net(w/o unlabeled) & - & - & 78.9 & 79.6 \\
I-Net & - & - & 79.5 & 79.9 \\
\hline
\end{tabular}
\end{center}
\end{table}

\begin{table}
\label{Table:ap}
\begin{center}
\caption{Comparisons between I-Net and OIM for detection.}
\begin{tabular}{|p{2cm}|p{2cm}|p{2cm}|}
\hline
Method & AP(\%) & Recall(\%) \\
\hline\hline
OIM & 74.9 & 79.1\\
\hline
I-Net & 79.6 & 82.2 \\
\hline
\end{tabular}
\end{center}
\end{table}

In experiment, the CMC top-1 accuracy and the mAP (mean average precision) are used for evaluating the person re-identification performance. The experiment with or without using unlabeled identities has been conducted, respectively. Specifically, the re-identification results are shown in Table 1, from which we can see that the proposed I-Net achieves a top-1 accuracy of 81.5\% and mAP of 79.5\% which outperforms all the compared methods. From Table 1 and Table 2, we observe that our I-Net outperforms the all single person re-identification methods with existing detectors, which demonstrate that it is important and necessary to integrate detection and re-id together for joint modeling. Note that, the GT denotes that the ground truth bounding boxes are directly used without further detection. Benefit from our siamese structure and real-time update OLP loss function, the feature stored in our model are fresher than OIM, which leads to a better result. On the other hand, our method outperforms NPSM by 1.6\% in mAP, and we have a lower computation cast, because they cascade several NPSM unit with parts of resnet50 to get the result.

Additionally, for better insight of the detection accuracy, the AP (average precision) and recall rate for OIM and the I-Net are measured. The results are shown in Table 2, from which we can see that the proposed I-Net shows significant superiority than OIM with 5\% improvement in AP and 3\% improvement in recall for pedestrian detection task with outperformed re-id accuracy. Therefore, our proposed method has good effect in both detection and re-id task.

In summary, the I-Net in end-to-end architecture keeps a dominated result in re-identification performance than the independent detection and re-id methods. The gap between person search and the real-world video surveillance application is further narrowed.

\subsection{Model Discussion}

\textbf{Analysis of Joint Loss.} Our HEP loss can be recognized as a variation of softmax loss, which treats the re-identification task as a classification problem. The difference between softmax loss and HEP loss is that the original softmax loss computes all identities (5533 classes) in the datasets, while our HEP loss computes only 100 classes for each subgroup. In fact, unlike the mini-batch sized network, we take 2 images as input of the I-Net. The number of identities from two images is much smaller than 5533, which makes a single softmax too hard to train. To have an insight of the co-training between OLP loss and HEP loss, 3 different cases have been discussed: OLP only, OLP with softmax loss, and OLP with HEP.

\begin{table}[h]
\label{Table:HEPsoft}
\begin{center}
\caption{Comparisons among different loss types}
\begin{tabular}{|m{4cm}|m{1.5cm}|m{1.5cm}|}
\hline
Loss Type& mAP(\%) & Top-1(\%)\\
\hline\hline
OLP only & 73.6 & 76.2 \\
\hline
OLP+softmax & 79.0 & 81.2 \\
\hline
OLP+HEP & 79.5 & 81.5 \\
\hline
\end{tabular}
\end{center}
\end{table}

The results with 3 types of loss functions have been shown in Table 3. We can see that single OLP loss achieves the worst result. By adding the softmax loss function in our framework, both the mAP and top-1 increase sharply because of the joint learning of verification and classification. Further, the HEP loss can be recognized as a special type of softmax by simultaneously considering hard example priority and randomness. Therefore, the joint loss of OLP+HEP shows the best re-identification result.

\textbf{Influence of Stored Features.} Another parameter that might influence the CNN model is the number of proposal features stored in the feature dictionary (i.e. dictionary size). In our implementation, this parameter is set as 40 times the mini-batch size. The number of proposals from each PPN is 128, therefore, $40\times128=5120$ features will be stored in the feature dictionary. To explore the influence of size of feature dictionary. OLP loss-only and the joint loss have been tested. The result is shown in Table 5. Large dictionary makes the feature stored out of date, while a dictionary with little features leads the loss function satisfy the condition easily. The best performance is achieved when joint loss of OLP+HEP with a dictionary size of 5120 is used.

\begin{table}[h]
\label{Table:bg}
\begin{center}
\caption{Influence of the number of feature stored in OLP.}
\begin{tabular}{|c|c|c|c|}
\hline
 & \multicolumn{3}{c|}{Number of Features} \\
\hline
mAP(\%) & $20\times128$ & $40\times128$ & $80\times128$ \\
\hline
OLP only & 73.2 & 74.3 & 72.9 \\
\hline
OLP+HEP & 78.4 & 79.5 & 79.1 \\
\hline\hline
Top-1(\%) & $20\times128$ & $40\times128$ & $80\times128$ \\
\hline
OLP only & 76.0 & 77.5 & 75.6 \\
\hline
OLP+HEP & 80.7 & 81.5 & 81.1 \\
\hline
\end{tabular}
\end{center}
\end{table}

\begin{figure}[t]
\begin{center}
   \includegraphics[width=0.9\linewidth]{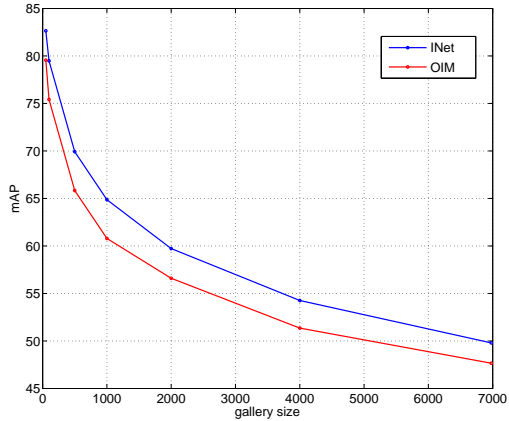}
\end{center}
   \caption{mAP changes with gallery size.}
\label{fig:gallerysize}
\end{figure}

\textbf{Gallery Size.} Person search problem should be more challenging when the gallery size is growing up. Therefore, we evaluate our method on different gallery size from 50 to 6978 (full set). All test images are covered even in a small gallery size. The result is shown in Fig.\ref{fig:gallerysize}. As the gallery size is increased, the model suffer a significantly descend in mAP. It means that more hard samples can be chosen alone with the increase of the gallery size in the test phase, which leads to a more difficult mission. Our method win about 2-3\% to the OIM in each gallery size.

\section{Conclusion}

In this paper, we introduce a novel end-to-end learning framework for large-scale person search mission in a whole image scene. By jointly modeling pedestrian detection and re-identification, an integrated convolutional neural network (I-Net) is proposed, which has structured a Siamese net architecture. Specifically, a novel on-line pairing loss (OLP) and hard example priority based softmax loss (HEP) are proposed for supervising the training of the person identification network. For joint loss computation, we further propose to design a feature dictionary which is used to store a large amounts of features, such that more negative pairs can be obtained to improve the training effect. HEP treats the re-identification task as a classification problem and prefers to handle the hard classes, which has improved the effectiveness as well as the efficiency. By jointly learning the I-Net end-to-end on the CUHK-SYSU dataset\cite{Xiao2016End}, the proposed model outperforms state-of-art in pedestrian re-identification as well as person detection.

\appendix

\bibliographystyle{named}
\bibliography{ijcai18}

\end{document}